\documentclass[letterpaper, 10 pt, journal, twoside]{IEEEtran}

\IEEEoverridecommandlockouts                              

\usepackage{array}
\usepackage{float} 
\usepackage{subfigure}
\usepackage{url}
\usepackage{verbatim}
\usepackage{graphicx}
\usepackage{cite}
\usepackage{booktabs}
\usepackage{makecell} 
\usepackage{multirow} 
\usepackage{amsmath,amssymb,amsfonts}
\usepackage{threeparttable}
\usepackage{bm}  
\usepackage[ruled]{algorithm2e}  
\usepackage[colorlinks,linkcolor=blue,citecolor=black]{hyperref}
\usepackage{balance}
\usepackage{xcolor}
\graphicspath{{FigV2/}}
\newcommand{\textquote}[1]{`{#1}'}
\definecolor{darkgreen}{RGB}{0,176,80} 
\title{\LARGE \bf
	Marker or Markerless? Mode-Switchable Optical Tactile Sensing for Diverse Robot Tasks
}

\author{Ni Ou$^{1,2,\dagger}$, Zhuo Chen$^{2}$ and Shan Luo$^{2}$
	\thanks{*This work was supported by the National Natural Science Foundation of China under Grant 62173038 and the EPSRC project ``ViTac: Visual-Tactile Synergy for Handling Flexible Materials" (EP/T033517/2).}
        \thanks{$^{1}$Ni Ou is with the State Key Laboratory of Intelligent Control and Decision of Complex Systems, Beijing Institute of Technology, Beijing, 100081, China.}
	\thanks{$^{2}$Ni Ou, Zhuo Chen and Shan Luo are with the Robot Perception Lab, Centre for Robotics Research, Department of Engineering, King's College London, London WC2R 2LS, United Kingdom. Emails: {\tt\small \{ni.ou, zhuo.7.chen, shan.luo\}@kcl.ac.uk}.}
	\thanks{$^{\dagger}$Work was done while Ni Ou was visiting King's College London.}
}
\begin{document}
	\maketitle
	\thispagestyle{empty}
	\pagestyle{empty}
	\begin{abstract}
		Optical tactile sensors play a pivotal role in robot perception and manipulation tasks. The membrane of these sensors can be painted with markers or remain markerless, enabling them to function in either marker or markerless mode. However, this uni-modal selection means the sensor is only suitable for either manipulation or perception tasks. While markers are vital for manipulation, they can also obstruct the camera, thereby impeding perception. The dilemma of selecting between marker and markerless modes presents a significant obstacle. To address this issue, we propose a novel mode-switchable optical tactile sensing approach that facilitates transitions between the two modes. The marker-to-markerless transition is achieved through a generative model, whereas its inverse transition is realized using a sparsely supervised regressive model. Our approach allows a single-mode optical sensor to operate effectively in both marker and markerless modes without the need for additional hardware, making it well-suited for both perception and manipulation tasks. Extensive experiments validate the effectiveness of our method. For perception tasks, our approach decreases the number of categories that include misclassified samples by 2 and improves contact area segmentation IoU by 3.53\%. For manipulation tasks, our method attains a high success rate of 92.59\% in slip detection. Code, dataset and demo videos are available at the project website \href{https://gitouni.github.io/Marker-Markerless-Transition/}{https://gitouni.github.io/Marker-Markerless-Transition/}
	\end{abstract}
	
	
	\section{Introduction}
	\label{Sec.Intro}
Tactile feedback offers valuable contact information for robot actuators, providing insights into the shape and texture of touched objects, as well as the contact forces, deformations and slip information~\cite{luo2017-tactile-perception-review}. In recent years, the development of vision-based tactile sensors, such as GelSight~\cite{Gelsight}, GelSlim~\cite{Gelslim}, and TacTip~\cite{Tactip-family}, has enabled robots to utilize high-resolution tactile images to facilitate challenging perception and manipulation tasks. During the fabrication of these sensors, markers are often applied to their elastomer for external force estimation, which is crucial for tasks like robotic manipulation~\cite{Gelsight,Gelslim}. Each marker's displacement is approximately proportional to the local applied force. The recovered force distribution from marker motions enables the optical tactile sensors with capabilities of analyzing shear~\cite{Measure-shear-slip}, estimating contact force~\cite{Gelslim-V2}, and detecting slip~\cite{improved-gelsight}. These capabilities are essential for robot manipulation and closed-loop control~\cite{Gripper-control1, Gripper-control2}.
	
Nevertheless, the presence of markers decreases the continuity of the tactile images and reduces the efficient perception area of the camera, thus inevitably disrupts the performance of robot perception in tasks like texture classification. Although these marker-overlaid tactile images can be classified by deep learning models~\cite{Active-clothing-perception, Spatio-temporal-attention}, other downstream tasks that require dense correspondences across tactile frames such as image stitching and segmentation~\cite{GelFinger}, remain challenging since these opaque markers disrupt feature extraction and matching.
	
	\begin{figure}[!t]
		\begin{center}
		  \includegraphics[width=0.98\linewidth]{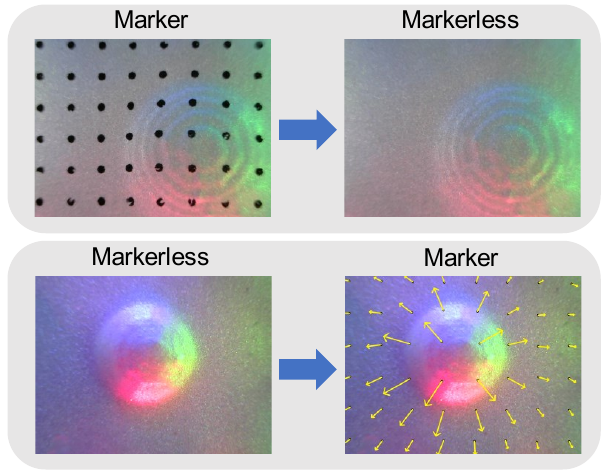}
		\end{center}
		\caption{Bidirectional transitions between marker and markerless modes. Marker-markerless transition (top): black markers are replaced with photo-realistic pixels; markerless-marker transition (bottom): pseudo marker motions (yellow arrows) are generated from a markerless tactile image.}
		\label{Fig.abstract}
	\end{figure}

Some researchers manage to deal with this conflicting problem from the perspective of hardware. One popular approach substitute conventional opaque markers with transparent ultraviolet (UV) markers~\cite{Low-cost-GelSight-UV,UVtac}. These markers that are made of UV ink appear transparent, and become illuminated when exposed to UV light. This type of sensor can switch between two modes: LED mode for tactile perception and UV LED mode for force field estimation. However, the mode-switch operation is controlled by toggling the UV or white LED lights, thereby complicating the design of the electrical circuit. Additionally, the illumination of UV markers is not instantaneous and the delay potentially introduces temporal inconsistencies between two modes.
 
Unlike the aforementioned studies, we propose a novel mode-switchable method to implement bidirectional transitions between marker and markerless status at the software level, with the function of our method illustrated in Fig.~\ref{Fig.abstract}. On the one hand, we utilize a generative model for marker-markerless transition. It transfers tactile images with markers into pseudo markerless tactile images, in which markers are replaced with photo-realistic RGB pixels. On the other hand, a sparsely supervised regressive model is employed to realize the markerless-marker transition, which involves placing pseudo markers on a markerless image. As a result, our method enables a single sensor to possess dual modes, allowing it to perform both robot perception and manipulation tasks. 

The key contributions of this paper are summarized below:
	\begin{itemize}
		\item We propose a mode-switchable optical tactile sensing method without need of any additional hardware, which enables bidirectional transitions between marker and markerless modes and allows a single-mode tactile sensor to perform diverse robot tasks;
		\item A novel diffusion-based framework is proposed for the marker-markerless transition, with a marker-offset strategy devised to make it adaptive to new sensors;
		\item Extensive experiments are conducted to verify the effectiveness of our method. Results show that our method exhibits high performance in both marker-markerless and markerless-marker transitions. 
	\end{itemize} 
	
	The remainder of this paper are organized as follows: Section~\ref{Sec.Related_Works} investigates recent related works to our study; Section~\ref{Sec.Method} describes our mode-switchable tactile sensing method; Section~\ref{Sec.Experiments} showcases a series of extensive experiments that validate the effectiveness of our method in various robot tasks; Section~\ref{Sec.Conclusion} concludes the paper and presents future research.
	
	\section{Related Works}
	\label{Sec.Related_Works}
	\subsection{Optical Tactile Sensors}
	\label{Subsec.Optical_Tactile_Sensors}
	Optical tactile sensors employ vision sensors and auxiliary light sources to capture detailed tactile information. Over last decades, researchers have developed a wide range of optical tactile sensors that can be broadly categorized into two groups: the TacTip family~\cite{Tactip-family} and the GelSight family~\cite{Gelsight}. The TacTip sensors~\cite{Tactip-original} comprise a black hemisphere membrane with white pins embedded in the tips, along with LED lights and a CCD camera. By mounting the camera to track the pin array, these sensors enable the measurement of applied force through marker movement. The original TacTip design has also been expanded with variants such as TacCylinder~\cite{TacCylinder} and TacTip-M2~\cite{TacThumb} to cater to specific applications.
	
	On the other hand, GelSight sensors~\cite{GelSight-original} utilize a transparent gel material covered with a top reflective layer. By utilizing light sources of different colors, the camera can capture deformation and geometry information through colored illumination, which can also be leveraged for depth reconstruction~\cite{Gelsight}. Furthermore, when the GelSight's elastomer is overlaid with markers, this sensor gains the ability to estimate external forces~\cite{Gelslim-V2} and detect slip~\cite{improved-gelsight}. Overall, GelSight sensors offer versatility in both perception and manipulation tasks. In this paper, we verify the effectiveness of our method on cube-shaped GelSight sensors, however, it can be adapted to other optical tactile sensors like GelTip~\cite{GelTip} for perception and manipulation tasks.
	
	\subsection{Generative Models for Optical Tactile Sensing}
	\label{Subsec.Generative_Models_Tactile_Sensing}
	Deep generative models exhibit high performance in synthesizing realistic images. They can be classified into two branches: Generative-Adversarial Networks (GANs) and likelihood-based methods. GANs have dominated the field of image generation for several years~\cite{GAN-original}. Their frameworks include a generator and a discriminator that are jointly trained under an adversarial strategy. Comparatively, likelihood-based methods encompass various sub-branches, such as variational autoencoders~\cite{VAE-original}, autoregressive models~\cite{Deep-autoregressive-Networks}, normalizing flows~\cite{Normalizing-Flows1} and diffusion models~\cite{DDPM}. 
	
	Generative models have been employed in tactile image generation to address the need for a large volume of tactile images in data-driven approaches. For example, GANs have been utilized to generate realistic tactile images, which are labor-intensive to collect in the real world~\cite{Sim-to-real-CycleGAN, Sim-to-real-ACTNet,Sim-to-real-texture-generation}. GANs have also been applied to produce tactile outputs with the prompts of visual images~\cite{Touching-to-see} or to produce haptic rendering from visual inputs~\cite{cao2023vis2hap}. Additionally, some researchers have leveraged a masked autoencoder for tactile image completion~\cite{TacMAE}, and demonstrated the effectiveness of generative models in reconstructing missing tactile signals. In this study, we leverage a diffusion-based generative model~\cite{DDPM} for our marker-markerless transition, as it has demonstrated superiority over GANs in various image editing tasks~\cite{Palette}.

  \begin{figure*}[!t]
		\centering
		\begin{center}
		\includegraphics[width=0.98\linewidth]{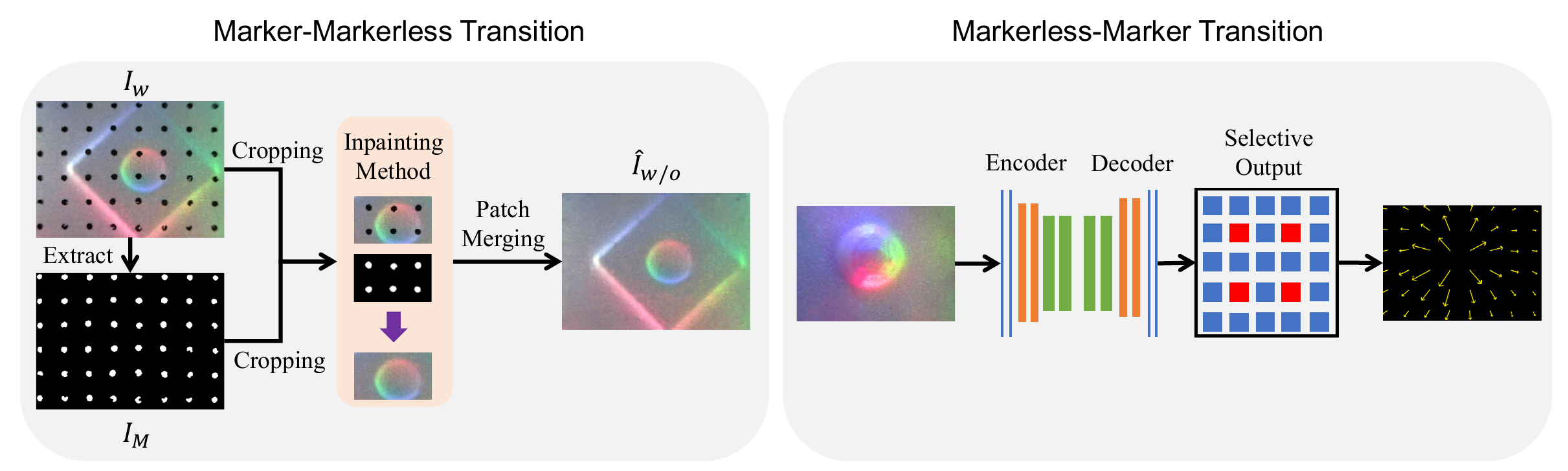}
		\end{center}
		\caption{Pipelines of the two transitions in our mode-switchable approach. The marker-markerless transition (left) is implemented by an inpainting method, with the tactile image with markers $I_w$ and the mask of markers $I_M$ as inputs. The markerless-marker transition (right) is realized by a encoder-decoder network that generates pseudo marker motions $\hat{M}_v$ from a markerless tactile image $I_{w/o}$. The selective output module retrieves features of sparse pixels to output 2D marker motions. Yellow arrows show the orientation and magnitude of marker motions.}
		\label{Fig.framework}
	\end{figure*}
 
\section{Method}
\label{Sec.Method}
\subsection{Overview}
\label{Subsec.overview}
	
As illustrated in Fig.~\ref{Fig.framework}, the framework of our mode-switchable method consists of two transitions. The marker-markerless transition (detailed in Section~\ref{Subsec.marker-markerless}) starts with a tactile image $I_w$ of width $W$ and height $H$ with markers, and markers $I_M$ extracted from $I_w$ following the marker extraction approach in~\cite{UVtac}. Subsequently, $I_w$ and $I_M$ are used as the inputs for an inpainting method to generate a pseudo markerless tactile image $\hat{I}_{w/o}$, wherein the regions of markers are replaced with RGB pixels to maintain texture and color continuity of $I_w$. On the other hand, the markerless-marker transition (detailed in Section~\ref{Subsec.markerless-marker}) employs a regressive neural network to predict a pseudo marker motion field $\hat{M}_v$ from the input markerless image $I_{w/o}$. The centers of markers are predetermined in the image, and their motions are represented by 2D vectors. 

\subsection{Marker-Markerless Transition}
\label{Subsec.marker-markerless}
The core of our marker-markerless transition is the inpainting module. Inpainting indicates restoring the missing pixels in a designated region of an image to maintain the color and texture consistency, with the pixels outside the region unchanged. Mathematically, the output of the inpainting method $\dot{I}_{w/o}$ is rectified by $I_w$ within the region of $I_M$, yielding $\hat{I}_{w/o}$ , which can be formulated as:
\begin{equation}
    \hat{I}_{w/o} = I_M \cdot \dot{I}_{w/o} + (1-I_M) \cdot I_w
    \label{Eq.inpainting}
\end{equation}

Here, inspired by~\cite{DDPM}, we propose an iterative inpainting method called \textit{TacDiff} based on diffusion models. As outlined in Algorithm~\ref{algo.TacDiff}, TacDiff predicts $\hat{I}_{w/o}$ from $I_w$ and $I_M$ based on (\ref{Eq.inpainting}). It initializes a noisy image $\hat{I}_{w/o}^{(T)}$ by adding Gaussian noise $\epsilon\sim \mathcal{N}(0,1)$ to the $I_M$ region of $I_w$. As a result, the differences between $I_{w/o}$ and $I_w$ only exist in $I_M$. For each subsequent iteration $t$, a denoising function $D(\hat{I}_{w/o}^{(t)},t)$ estimates the noise-free image $\dot{I}_{w/o}^{(t-1)}$  from $\hat{I}_{w/o}^{(t)}$ , while a noise-adding function $A(\dot{I}_{w/o}^{(t-1)},t-1)$ adds $t-1$ level of noise to the $I_M$ region of $\dot{I}_{w/o}^{(t-1)}$, yielding a less noisy image $\hat{I}_{w/o}^{(t-1)}$. After $T$ iterations, we can obtain $\hat{I}_{w/o}^{(0)}$ as the final output for $\hat{I}_{w/o}$.
\begin{algorithm}[htbp]
	\caption{TacDiff}
	\label{algo.TacDiff}
	\KwIn{$I_w, I_M$}
	\KwOut{$\hat{I}_{w/o}$}
	Sample $\epsilon\sim \mathcal{N}(0,1)$ \\
	$\hat{I}_{w/o}^{(T)}=I_M\cdot \epsilon+(1-I_M)\cdot I_w$\\
	\For{$t=T,T-1,...,1$}
	{
		$\dot{I}_{w/o}^{(t-1)}=D(\hat{I}_{w/o}^{(t)},t)$\\
		$\hat{I}_{w/o}^{(t-1)}=I_M\cdot A(\dot{I}_{w/o}^{(t-1)},t-1)+(1-I_M) \cdot I_w$
	}
\end{algorithm}

As introduced in~\cite{DDPM}, the noise-adding function $A$ is predetermined while the denoising function $D$ is learned by a neural network. A U-Net~\cite{UNet} is employed as $\hat{D}_\theta(\hat{I}_{w/o}^{(t)},t)$ to function as $D$ and trained using the following loss:

\begin{equation}
    \label{Eq.TacDiff-loss}
    L_{dif}=\left\Vert I_M\cdot[\hat{D}_\theta(\hat{I}_{w/o}^{(t)},t) - I_{w/o}] \right\Vert_2^2
\end{equation}
\noindent where $\theta$ is the parameters of the network to be trained.

\begin{figure}[!t]
		\centering
		\includegraphics[width=0.9\linewidth]{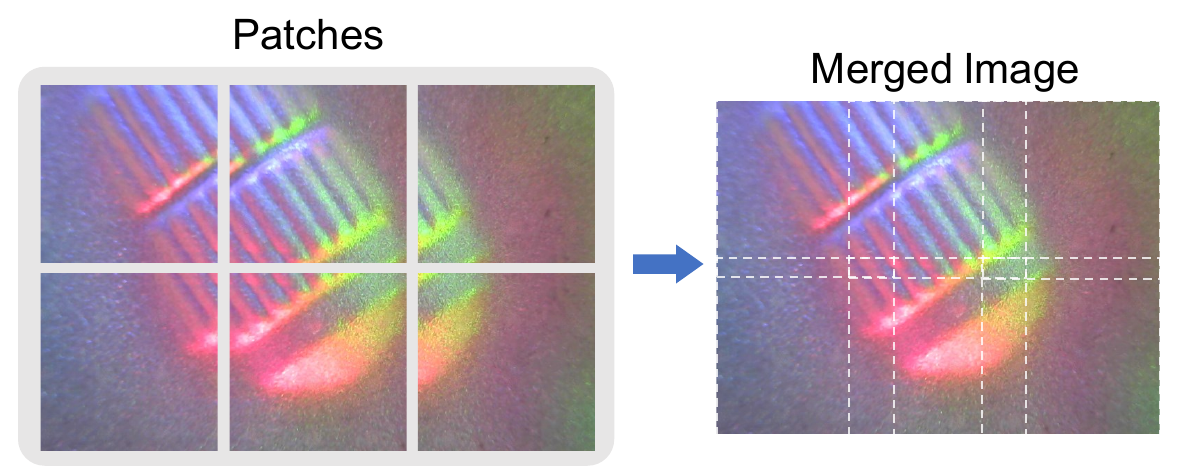}
		\caption{Patch-based Merging. The resulting image is obtained by merging six small patches that are separately predicted by the inpainting model.  The regions of patches in the merged image are annotated with white dashed rectangles, which overlap each other.}
		\label{Fig.patch_emerge}
	\end{figure}

\noindent \textbf{Patch-based Merging.} Due to the resolution limitation of U-Net, the final output of is synthesized by merging multiple low-resolution patches predicted by $\hat{D}_\theta$ using a weighted algorithm. For instance, as demonstrated in Fig.~\ref{Fig.patch_emerge}, to generate a 640x480 tactile image, the model predicts six 256x256 patches with overlapping regions. Pixels within the non-overlapping areas directly retain the values from their corresponding patches, while those within the overlapping areas achieve the average values of the patches that cover this area.

\begin{figure}[htbp]
    \centering
    \includegraphics[width=0.9\linewidth]{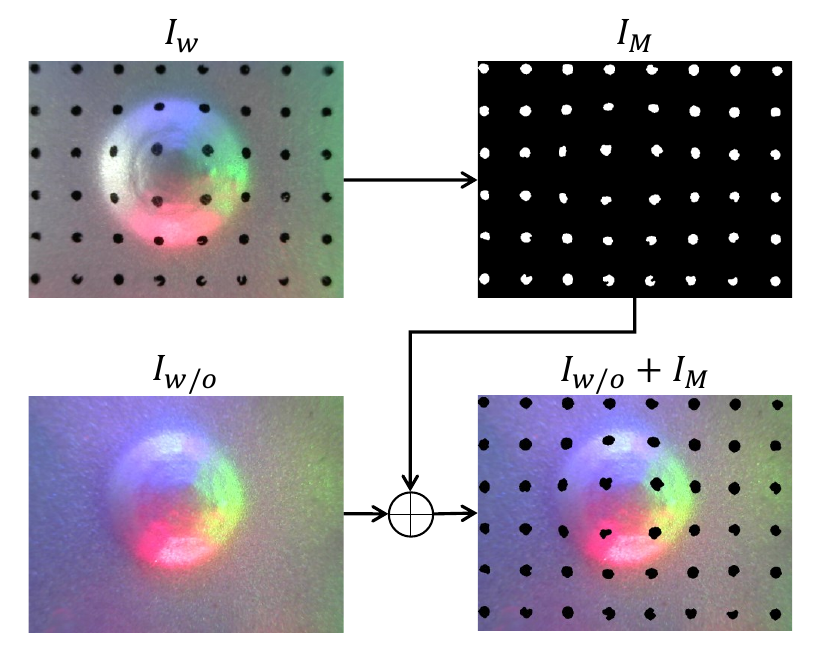}
    \caption{Training data acquisition for the marker-markerless transition. $I_M$ is extracted from $I_w$ and then placed onto $I_{w/o}$ to form the new tactile image with markers $I_{w/o}+I_M$. $I_{w/o}+I_M$ serves as input while $I_{w/o}$ serves as the ground-truth output.}
    \label{Fig.marker_markerless_data}
\end{figure}

\noindent \textbf{Training Data Acquisition.} We fabricate two GelSight sensors with the same tactile resolution and membrane material: one with markers (Sensor WM) and one without markers (Sensor WO). The object is pressed onto both Sensor WO and Sensor WM, respectively, at the same position and depth, yielding a pair of $I_{w/o}$ and $I_w$. Since inpainting methods require the background of $I_w$ and $I_{w/o}$ to be the same, we extract the markers of $I_w$, i.e., $I_M$, and then place $I_M$ onto $I_{w/o}$ to satisfy this condition. In the case of the marker-markerless transition, $I_{w/o}+I_M$ and $I_{w/o}$ serve as input and output, respectively.

\begin{figure}[htbp]
		\centering
		\includegraphics[width=0.98\linewidth]{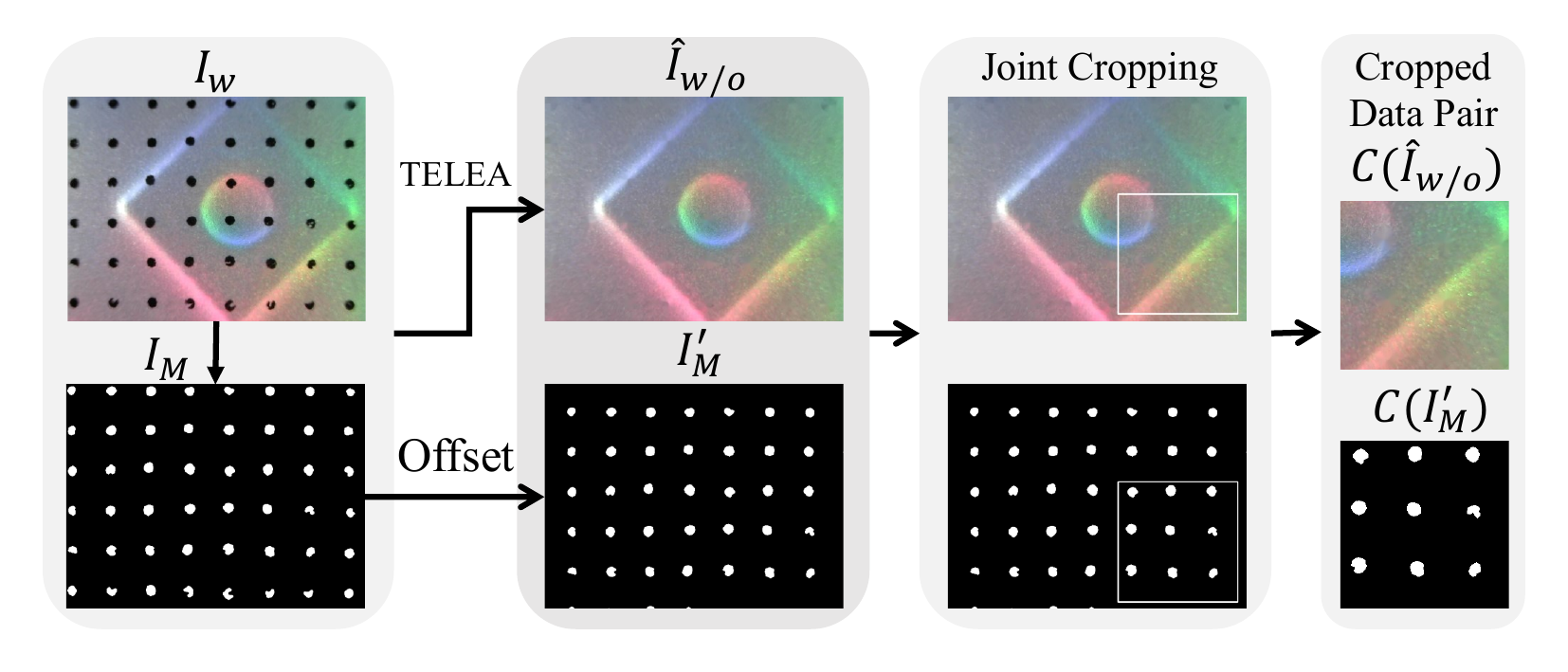}
		\caption{Marker-offset strategy for training TacDiff if Sensor WO is not available. The TELEA inpainting algorithm~\cite{TELEA} is applied to generate $\hat{I}_{w/o}$. $I_M$ is translated by an offset to get $I'_M$. $\hat{I}_{w/o}$ and $I'_M$ are jointly cropped to obtain a pair of cropped patches $C(\hat{I}_{w/o})$ and $C(I'_M)$ for training TacDiff.}
		\label{Fig.zero_shot_finetune}
	\end{figure} 
    
    \noindent \textbf{Marker-offset Strategy.} The aforementioned pipeline is only applicable to the scenarios where paired $I_w$ and $I_{w/o}$ are available. When users utilize their own Sensor WM to collect $I_w$ and do not have a matched Sensor WO to collect $I_{w/o}$, we also develop a marker-offset strategy to train TacDiff.  As illustrated in Fig.~\ref{Fig.zero_shot_finetune}, this strategy enables users to finetune our TacDiff in their own dataset, eliminating the need for fabricating Sensor WO. Initially, $I_M$ is extracted from the original tactile image $I_w$. Next, a non-data-driven inpainting algorithm, such as TELEA \cite{TELEA}, is employed to generate a pseudo markerless image $\hat{I}_{w/o}$  with $I_w$ and $I_M$ as inputs. Meanwhile, a new mask $I'_M$ is created through translating the positions of markers by a constant offset ($\Delta x, \Delta y$):
    \begin{equation}
    \label{Eq.marker_offset}
    I'_M(i+\Delta x, j+\Delta y)=1,\; s.t.\;
        \begin{cases}
            I_M(i,j)=1 \\
            0\leq i+\Delta x<W \\
            0\leq j+\Delta y<H
        \end{cases}
    \end{equation}
    and
    \begin{equation}
    \label{Eq.marker_exclusion}
    I'_M(i,j) = 0,\;s.t.\;I_M(i,j)=1,
    \end{equation}
    to ensure that $I_M\cap I'_M=\varnothing$. Empirically, for evenly distributed markers like those shown in Fig.~\ref{Fig.zero_shot_finetune}, $\Delta x$ and $\Delta y$ are set to half horizontal and vertical distance between markers. Finally, a joint cropping operation is performed on $\hat{I}_{w/o}$ and $I'_M$ to obtain cropped markerless image $C(\hat{I}_{w/o})$ and corresponding mask $C(I'_M)$ for training TacDiff,then the updated loss function is formulated as:
    \begin{equation}
    \label{Eq.loss_offset}
    L_{dif}^*=\left\Vert C(I'_M)\cdot[\hat{D}_\theta(\hat{I}_{w/o}^{(t)},t) - C(\hat{I}_{w/o})] \right\Vert_2^2
    \end{equation}
    where TacDiff is trained on surrounding RGB pixels rather than those generated by TELEA. The cropping operation force the resolution of patches to match the input resolution of U-Net and augment the positions of markers to avoid overfitting as well.

\noindent \textbf{Baselines.} We also include two non-data-driven inpainting methods in our experiments for comparison, i.e., NS~\cite{NS} and TELEA~\cite{TELEA}. The NS method draws inspiration from fluid dynamics and utilizes a vector field defined by the stream function to transport the Laplacian of the image intensity into the inpainting region. In contrast, the TELEA method estimates the smoothness of unknown pixels by computing a weighted average over neighboring known pixels. In this approach, the missing regions are treated as level sets, and image information is propagated for inpainting using a Fast Marching Method (FMM)~\cite{FMM}.

\subsection{Markerless-Marker Transition}
\label{Subsec.markerless-marker}
As shown in the right half of Fig.~\ref{Fig.framework}, the regressive network predicts displacements of markers $\hat{M}_v$ from a markerless image $I_{w/o}$. In terms of implementation, the center coordinates of these markers $M_c$ are predetermined in $I_M$, allowing the model to output a 2D vector ($\delta x_i,\delta y_i$) for each marker center $i$. Our regressive network is modified from Encoder-Decoder network named DeeplabV3~\cite{DeepLabV3}. We substitute its cross-entropy loss with Mean Squared Error (MSE) loss to make it applicable to this regression task. In this case, since we only have ground-truth ($\delta x_i,\delta y_i$) at $M_c$ ($M_v$), we exclusively back-propagate gradients with respect to $M_v$, meaning that DeeplabV3 is sparsely supervised.

\noindent \textbf{Marker Motions vs. Marker Generation.} There are three reasons behind our choice to solely output marker motions $M_v$ instead of generating the binary mask of markers $\hat{I}_M$. First, in robotic grasping with optical tactile sensors~\cite{improved-gelsight, GelWedge, Gelslim}, the motions of markers rather than the markers themselves are used for force estimation and slip detection. Second, compared to generating marker shapes directly, predicting marker center motions is a sparse task that requires fewer computational resources for training. Third, the artifacts in $\hat{I}_M$ can be incorrectly recognized as markers and result in the failure of marker tracking, while directly predicting marker motions can avoid this problem. 

 \begin{figure}[t]
     \centering
     \includegraphics[width=0.9\linewidth]{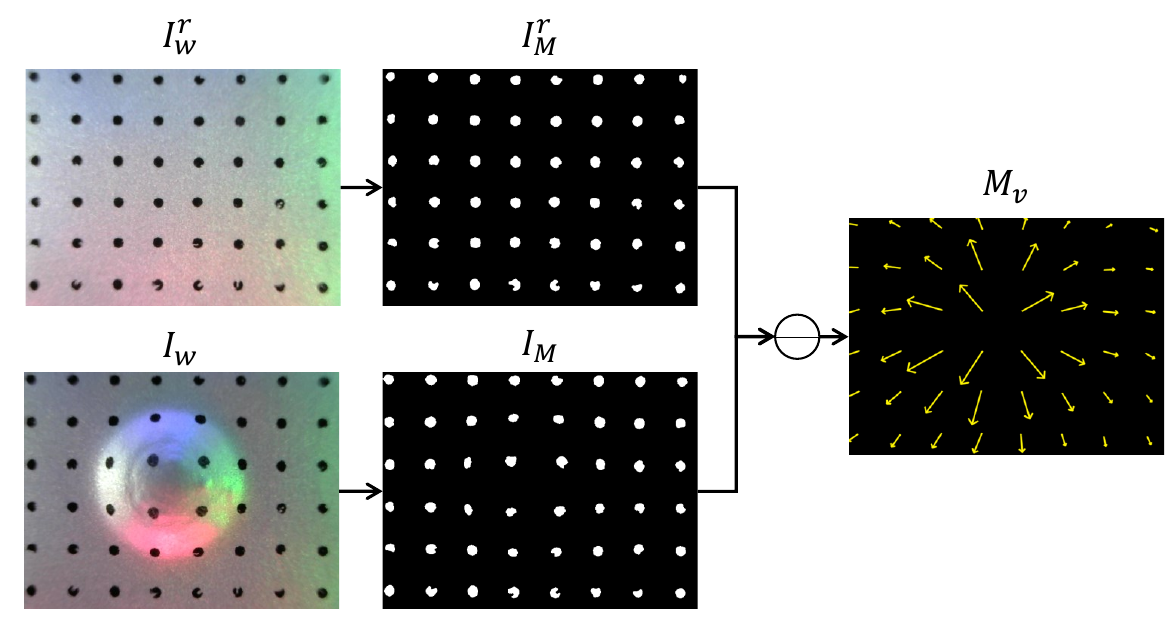}
     \caption{Training data acquisition for the markerless-marker transition. $I_w$ and $I_w^r$  (reference) are collected from Sensor WM, and $I_M$ and $I_M^r$ are their respective markers. Yellow arrows in $M_v$ represent the marker motions obtained from $I_M$ and $I_M^r$.}
     \label{Fig.markerless_marker_data}
 \end{figure}

\begin{figure*}[!t]
    \begin{center}
    \includegraphics[width=0.98\linewidth]{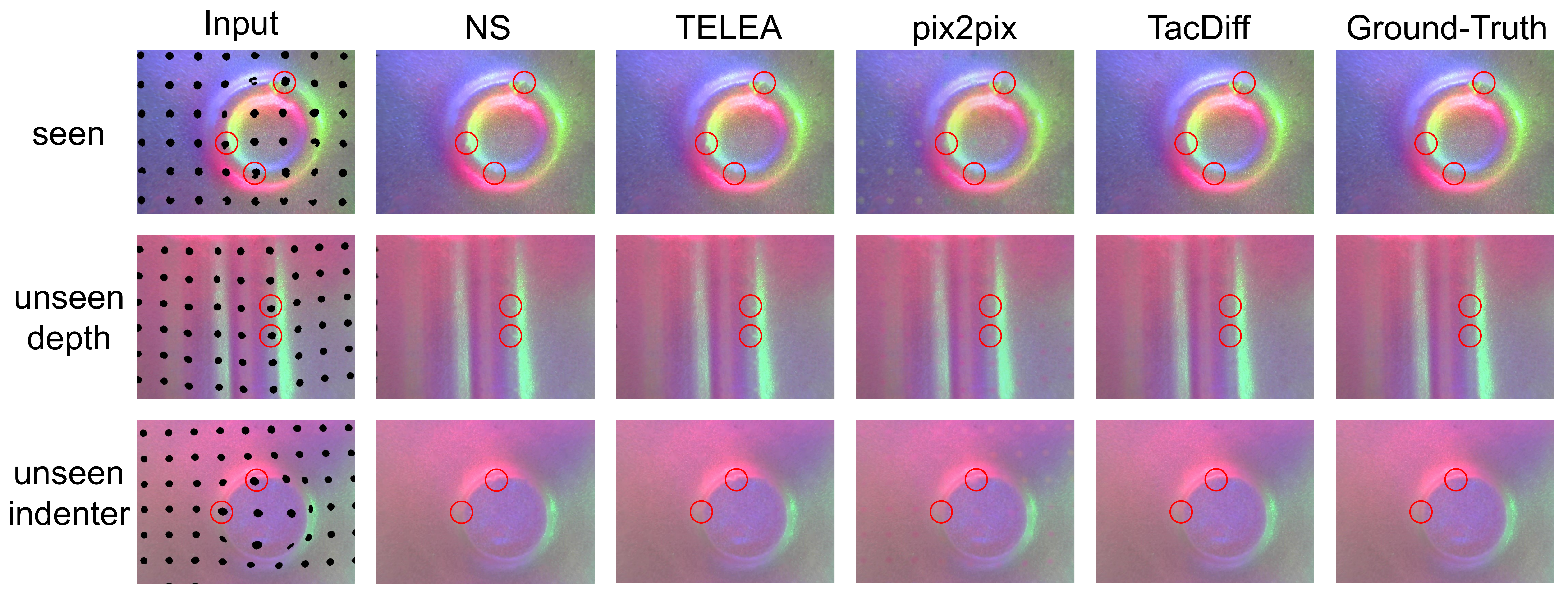}
    \end{center}
    \caption{Qualitative marker-markerless assessment. Some significant differences across tactile images are annotated with red circles. The leftmost circle in the 1$^\textup{st}$ row demonstrates that NS, TELEA, and pix2pix produce inconsistent textures on the ring, whereas TacDiff does not. Similar patterns can be observed in the three circles annotated in the tactile images in the 2$^\textup{nd}$ row.}
    \label{Fig.imq_res}
\end{figure*}

\noindent \textbf{Training Data Acquisition.} The approach to acquiring training data for our sparsely supervised marker motion prediction is presented in Fig.~\ref{Fig.markerless_marker_data}. First, a reference tactile image without any contact $I_w^r$ and a tactile image with contact $I_w$ are collected from Sensor WM. Ground-truth $M_v$ is obtained by subtracting marker positions in $I_M$ and $I_M^r$, where marker correspondences are built through nearest neighbor searching.

\noindent \textbf{Slip detection.} To apply this markerless-marker transition to robot manipulation tasks, we also devise a slip detection algorithm adapted from~\cite{improved-gelsight}: 
if the maximum marker motion between the current and the first collected tactile image exceeds a predetermined threshold $\epsilon_v$, we assume the slip happens:
 \begin{equation}
     \max_{(x_i, y_i\in M_c}\sqrt{\left( x_i - x_i^{(0)}\right)^2 + \left( y_i - y_i^{(0)}\right)^2} > \epsilon_v
     \label{Eq.slip-condition}
 \end{equation}
where $(x_i, y_i)$ and $(x_i^{(0)}, y_i^{(0)})$ are the locations of the $i^\textup{th}$ marker in the current and the first collected tactile images, respectively.

\section{Experiments}
\label{Sec.Experiments}

This section presents a series of experiments to evaluate the effectiveness of our method. We conduct upstream experiments to evaluate the image quality of pseudo markerless images and the precision of pseudo marker motions, as well as downstream experiments that involve tactile recognition and manipulation tasks to validate our marker-markerless and markerless-marker transitions, respectively. We also include a baseline method pix2pix~\cite{pix2pix} for comparison, which is designed for image-to-image translation. For the marker-markerless transition, the input and output of pix2pix are $I_{w/o}+I_M$ and $\hat{I}_{w/o}$, respectively. For the markerless-marker transition, $I_{w/o}$ serves as its input while the estimated mask of markers $\hat{I}_{M}$ serves as its output.

 \subsection{Dataset Description}
 \label{Subsec.dataset_descript}
\noindent \textbf{Paired Dataset A.}
To provide ground-truth tactile images in mode transitions, pairs of $I_{w/o}$ and $I_w$ are needed. Following prior works~\cite{Daniel-sim2real}, we use indenters to press against the sensors to obtain tactile images. The indenters used include 21 3D printed objects from~\cite{Daniel-sim2real} and 4 daily objects with richer textures compared to primitive geometric shapes in~\cite{Daniel-sim2real}: a sandpaper, a star-shaped screw driver bit, a hexagon-shaped screw driver bit, and a screw driver (whose handle was used for data collection). These objects were vertically pressed onto Sensor WM and WO at the same position and depth (1mm) to obtain pairs of $I_w$ and $I_{w/o}$. This process is controlled by a UR5e robot, whose position precision is $\pm$0.03 mm. In total, there are 128 pairs of tactile images in the dataset, and the contact area of each sample is manually annotated for the segmentation task. The dataset was randomly divided into training and testing sets with a ratio of 85:15.

\noindent \textbf{Paired Dataset B.}

The data collection methodology employed for this dataset is akin to that of Paired Dataset A, differing in marker patterns, light conditions of sensors, and the inclusion of additional shear motions of indenters. This dataset contains 2320 pairs of $I_{w/o}$ and $I_w$ recorded at the depth of 0.5mm, 1mm and 1.5mm. Due to significant labor costs, data collection was limited to 8 objects from~\cite{Daniel-sim2real}. Paired Dataset B was partitioned to assess the generalization capabilities of our approach on unseen indenters and contact depths. For unseen indenter experiments, our models are trained on data from 6 indenters and evaluated on data from the other 2. For unseen depth experiments, the training data contain depths of 0.5mm and 1.5mm, while the test data include depth of 1.0mm.

\noindent \textbf{ViTac.}
ViTac is a visual-tactile dataset \cite{Vitac} of 24 classes of garments with different textures. We only use its tactile modality, which contains tactile images with markers of different textures. Notably, we excluded empty tactile images without any contact using a imaging high-pass filter, and sampled 720 samples (30 for each class) among the remaining data. The dataset was randomly divided into training and testing sets with a ratio of 0.5:0.5.

\noindent \textbf{Slip Dataset.}
As illustrated in the first row of Fig.~\ref{Fig.grasp-vitac}, we controlled a robot gripper to hold an object and applied manual external force to induce slippage. We only collected tactile images from the Sensor WO used in Paired Dataset A, and the serially collected tactile images cover the entire process before and after the sliding occurs. In total, we collected 12 sequences where no slip occurred and 15 sequences where slip occurred for slip detection.
\subsection{Upstream Experiments}
\label{Subsec.upstream_ex}
\subsubsection{Marker-markerless}
\label{Subsec.up_ex_marker_markerless}

Regarding the marker-markerless transition, we conduct experiments on Paired Dataset A \& B to evaluate the similarity between the pseudo markerless tactile images and the ground truth tactile images. As shown in Fig.~\ref{Fig.imq_res}, pix2pix generates tactile images with artifacts at marker positions, due to its attempt to generate the entire image instead of specific regions. In contrast, TacDiff yields visually superior results compared to the other methods, particularly noticeable at the edges of the in-contact object.

\begin{table}[!t]
\caption{Image Quality Assessment}
\centering
    \begin{tabular}{cc|ccccc}
        \toprule
        ~ &~ &FID$\downarrow$ &KID$\downarrow$ &MSE$\downarrow$ &SSIM$\uparrow$  &PSNR$\uparrow$ \\
        \midrule
         \multirow{2}{*}{seen} 
         &NS &25.98 &2.08E-2 &4.104 &\textbf{0.982}&42.41 \\
        ~ &TELEA &30.94 &2.73E-2 &\textbf{3.469} &0.982&\textbf{42.84} \\
        ~ &pix2pix &142.4 &1.40E-1 &30.77 &0.914 &33.32 \\
        ~ &TacDiff &\textbf{2.169} &\textbf{1.12E-4} &3.780 &0.978 &42.48 \\
        \midrule
        \multirow{2}{*}{\makecell[c]{unseen \\depth}} 
        &NS &20.57 &1.54E-2 &4.022 &\textbf{0.987} &43.18 \\
        ~ &TELEA &23.37 &2.15E-2 &\textbf{2.268} &0.986&\textbf{44.76} \\
        ~ &pix2pix &62.45 &6.12E-2 &4.901 &0.974 &41.42 \\
        ~ &TacDiff &\textbf{3.216} &\textbf{2.48E-3} &3.147 &0.981 &43.27 \\
        \midrule
        \multirow{2}{*}{\makecell[c]{unseen \\indenter}} 
        &NS &20.86 &1.73E-2 &3.517 &\textbf{0.987}&43.96 \\
        ~ &TELEA &25.41 &2.50E-2 &\textbf{1.976} &0.986&\textbf{45.31} \\
        ~ &pix2pix &59.31 &5.78E-2 &5.328 &0.975 &41.23 \\
        ~ &TacDiff &\textbf{3.186} &\textbf{2.57E-3} &2.387 &0.980 &44.40 \\
        \bottomrule
    \end{tabular}
    \label{Table.img_qual}
\end{table}

In line with previous works~\cite{Daniel-sim2real}, we employed five metrics to quantitatively evaluate the image quality among different methods: Frechet Inception Distance (FID), Kernel Inception Distance (KID), Mean Squared Error (MSE), Structural Similarity Index Measure (SSIM), and Peak Signal-to-Noise Ratio (PSNR). As demonstrated in Table~\ref{Table.img_qual}, all three inpainting methods, i.e., TacDiff, NS and TELEA, outperform pix2pix, showing the efficacy of the inpainting approach in the marker-markerless transition. Furthermore, TacDiff exhibits superior performance to NS and TELEA in terms of FID and KID, despite slightly lower scores in other three metrics. This phenomenon is also observed in other diffusion-based applications~\cite{Cold-diffusion}. FID and KID scores have been widely taken as more important metrics as they measure the distribution similarity while the others are pixel-wise metrics that are susceptible to noise in the tactile images. Furthermore, it is observed in the last two groups of Table~\ref{Table.img_qual} that TacDiff still achieves the best FID and KID and performs second only to TELEA in terms of MSE and PSNR on both unseen datasets, highlighting its strong generalization ability.

\begin{table}[!t]
    \caption{Marker Motion Accuracy}
    \centering
    \begin{tabular}{cc|ccccc}
    \toprule
        ~&~&\multicolumn{2}{c}{$e_\textup{rmse}\downarrow$}&\multicolumn{2}{c}{$e_\textup{mag}\downarrow$}&\multirow{2}{*}{Nan\%$\downarrow$} \\
        ~&~ &mean &median &mean &median &~ \\
    \midrule
    \multirow{2}{*}{seen} &pix2pix &3.372 &2.211 &2.079 &1.219 &5.26 \\
    ~    &Ours &\textbf{1.814} &\textbf{0.730} &\textbf{1.369} &\textbf{0.507} &\textbf{0} \\
    \midrule
    \multirow{2}{*}{\makecell[c]{unseen \\depth}} &pix2pix &5.253 &2.566 &0.959 &0.458 &41.85 \\
    ~    &Ours &\textbf{2.059} &\textbf{1.278} &\textbf{0.137} &\textbf{0.078} &\textbf{0} \\
    \midrule
    \multirow{2}{*}{\makecell[c]{unseen \\indenter}} &pix2pix &7.029 &4.004 &1.537 &1.536 &80.45 \\
    ~    &Ours &\textbf{2.598} &\textbf{1.668} &\textbf{0.214} &\textbf{0.124} &\textbf{0} \\
    \bottomrule
    \end{tabular}
    \label{Table.marker_motion_seen}
\end{table}

\begin{figure}[t]
    \begin{center}
    \includegraphics[width=0.98\linewidth]{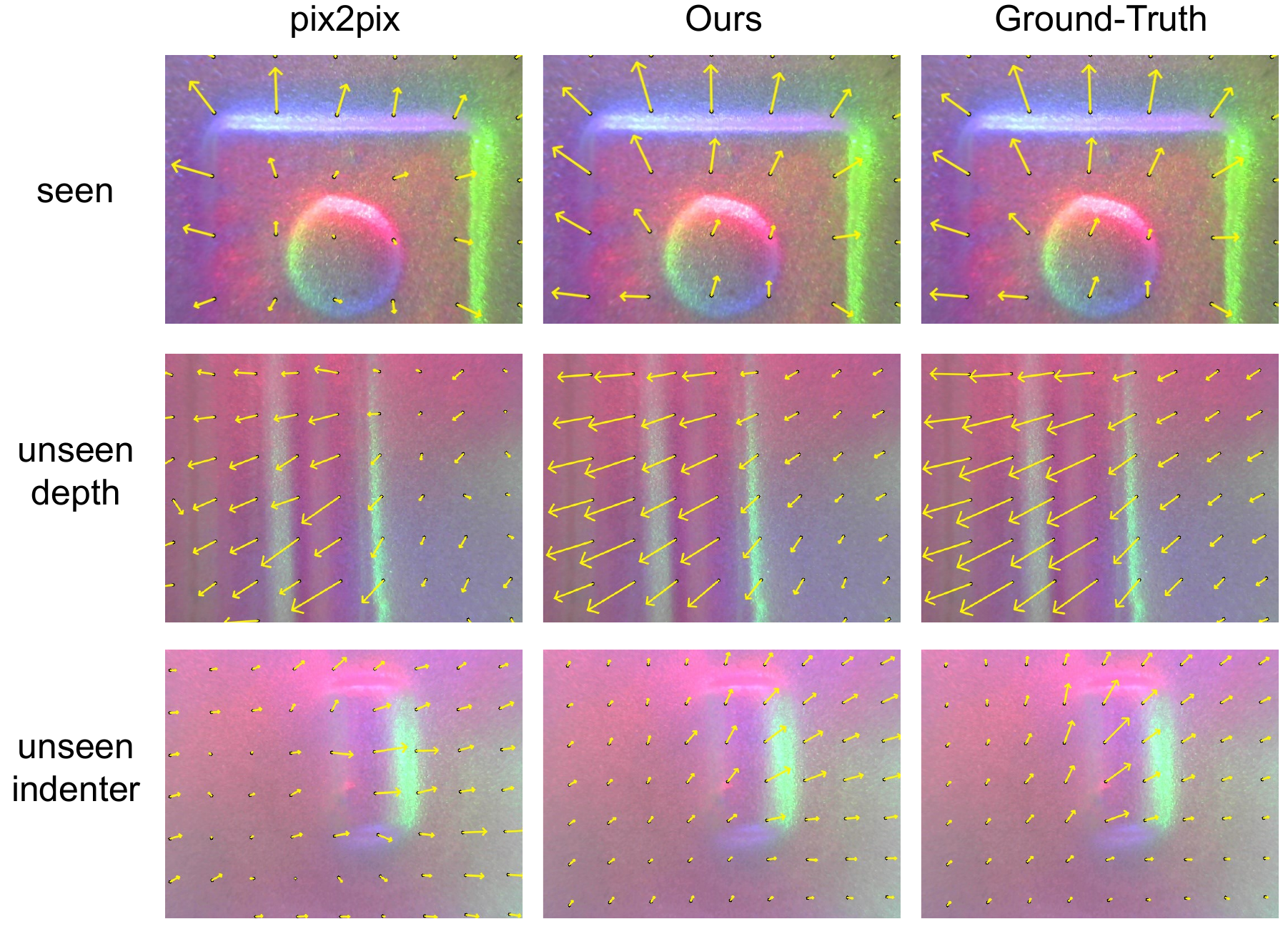}
    \end{center}
    \caption{Marker motion prediction. Figures in the $1^\textup{st}$ row depict normal indenter motion, while those in the $2^\textup{nd}$ and $3^\textup{rd}$ rows illustrate shear indenter motion}. The orientation and magnitude of marker motions predicted by our method are visually better than those predicted by pix2pix. For better visualization, the length of the arrows are six times the real magnitude of corresponding marker motions for the $1^\textup{st}$ row and two times for the $2^\textup{nd}$ and $3^\textup{rd}$ rows.
    \label{Fig.marker_motion_res}
\end{figure}

\subsubsection{Markerless-marker}
\label{Subsubsec.up_ex_markerless_marker}
	As the precision of marker motions reflects the efficacy of our markerless-marker transition, following~\cite{Marker-embedded-image-generation}, Root Mean Squared Error (RMSE) $e_\textup{rmse}$ and the magnitude error $e_\textup{mag}$ are employed to qualify the discrepancy between the ground-truth and predicted marker motions. Furthermore, as we discussed in Section~\ref{Subsec.markerless-marker}, pix2pix can generate artifacts in $\hat{I}_M$ so that the marker tracking process can fail in some cases. Two failure examples can be found in the  $2^\textup{nd}$ and $3^\textup{rd}$ rows of Fig.~\ref{Fig.marker_motion_res}, where the number of reference markers in \textquote{pix2pix} does not equal that in \textquote{Ground-Truth}. We only compute the metrics of pix2pix on success cases and record its proportion of failure with the metric Nan\%.

    As shown in the first row of Table~\ref{Table.marker_motion_seen}, both the average of $e_\textup{rmse}$ and the average of $e_\textup{mag}$ of our method are less than two-thirds of pix2pix's results while the median of $e_\textup{rmse}$ and the median of $e_\textup{mag}$ of our method are less than half of pix2pix's results. On unseen datasets, our method achieves almost 60\% lower $e_\textup{rmse}$ and $e_\textup{mag}$ compared with pix2pix, showing its generalization capability on contact depth and indenters. Notably, some results on unseen data are better than those on seen data because the quantity of training data of our Paired Dataset B is much larger than that of Paired Dataset A. In terms of qualitative results, Fig.~\ref{Fig.marker_motion_res} demonstrates that our method produces marker motion fields more closely aligned with ground truth than pix2pix.
\subsection{Downstream Experiments}
 \label{sec:downstream}
 \label{Subsec.downstream_ex}
    We follow prior works~\cite{UVtac, GelFinger, Measure-shear-slip} to carry out downstream experiments that show the significance of our method in tactile perception and manipulation tasks. The experiments include texture classification, contact area segmentation, slip detection and grasping tasks. The classification and segmentation experiment illustrates how the recovered intricate textures in our marker-markerless transition help tactile recognition, while slip detection and grasping tasks verify the effectiveness of our markerless-marker transition in predicting marker motions during grasping tasks.
\subsubsection{Classification and Segmentation}
\label{Subsubsec.segmentation}
We conduct texture classification and contact area segmentation experiments on ViTac Dataset and Paired Dataset A, respectively. With the marker-markerless transition, tactile images with markers are transformed into pseudo markerless tactile images for these two perception tasks. For comparison, we include a control group wherein tactile images with markers are directly utilized in perception tasks, designated as \textquote{none} in the subsequent discussion. Since we did not have ground-truth $I_{w/o}$,  we trained TacDiff and pix2pix using the marker-offset strategy introduced in Fig.~\ref{Fig.zero_shot_finetune}. We employ a ResNet-18~\cite{ResNet} network for classification and a DeepLabV3~\cite{DeepLabV3} network for segmentation.

\begin{table}[!t]
    \caption{Performance of Texture Classification}
    \centering
    \begin{threeparttable}
        \begin{tabular}{cc|cccc}
            \toprule
            ~ &none &NS &TELEA &pix2pix &TacDiff \\
            \midrule
            Acc $\uparrow$ &96.7\% &96.7\% &96.9\% &96.7\% &\textbf{97.5\%} \\
            Num\tnote{1} $\downarrow$ &7 &7 &7 &6 &\textbf{5} \\
            \bottomrule
        \end{tabular}
        \begin{tablenotes}
            \item[1] Number of classes containing misclassified samples.
        \end{tablenotes}
    \end{threeparttable}
    \label{Table.Classfication}
\end{table}
\begin{table}[!t]
    \caption{Performance of Contact Area Segmentation}
    \centering
    \begin{tabular}{cc|cccc}
        \toprule
        ~ &none &NS &TELEA &pix2pix &TacDiff \\
        \midrule
        Acc $\uparrow$ &98.25\% &98.22\% &98.42\% &98.43\% &\textbf{98.44\%} \\
        IoU $\uparrow$ &83.53\% &82.94\% &83.53\% &83.53\% &\textbf{87.06\%} \\
        \bottomrule
    \end{tabular}
    \label{Table.Segmentation}
\end{table}

 Regarding classification results, Table \ref{Table.Classfication} demonstrates that TacDiff achieves the best performance in all metrics. Compared with \textquote{none}, TacDiff improves the classification accuracy to by 0.8\% and reduces the misclassified classes by 2. Concerning segmentation segmentation tasks, it is shown in Table~\ref{Table.Segmentation} that our method improves 3.53\% IoU and 0.19\% accuracy compared with \textquote{none}.
 
\begin{figure*}[!t]
    \begin{center}
        \includegraphics[width=0.9\linewidth]{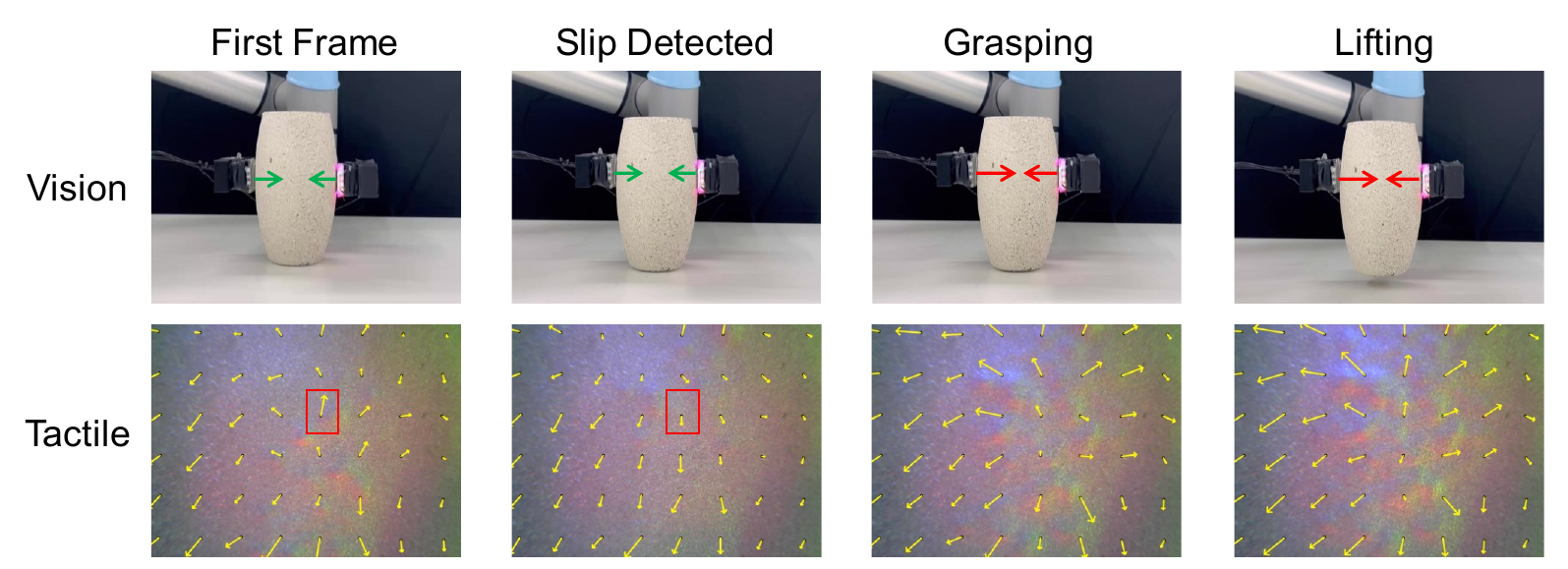}
    \end{center}
    \caption{An illustration of grasping with our slip detection feedback for the leftmost object in Fig.~\ref{Fig.grasp-obj-demo}. Initially, the robot attempts to lift a cup but encounters slip (2$^\textup{nd}$ column). The slip is detected using our algorithm and the grasping force is then increased (3$^\textup{rd}$ column), with the cup being lifted successfully thereafter (4$^\textup{th}$ column). Green arrows denote the initially applied grasping force, while red arrows indicate force adjustments. The marker motion triggering slip condition (\ref{Eq.slip-condition}) is highlighted with a red rectangle in the 1$^\textup{st}$ and 2$^\textup{nd}$ columns of tactile images. Arrow lengths are scaled 14 times for clarity. Visit our project website for a demo video.}
    \label{Fig.grasp-vitac}
\end{figure*}
\subsubsection{Slip detection and grasping}
 	
\begin{figure}[!t]
         \centering
    \includegraphics[width=0.9\linewidth]{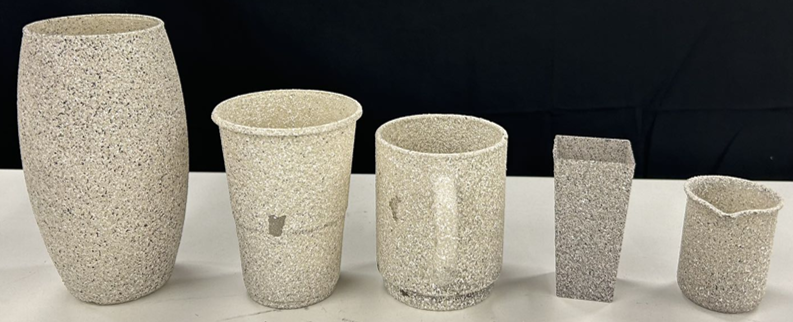}
         \caption{Objects used for grasping. They are numbered 1 to 5 from left to right in our paper.}
         \label{Fig.grasp-obj-demo}
     \end{figure}

Our slip detection experiment is carried out on Slip Dataset using the slip detection algorithm detailed in Section~\ref{Subsec.markerless-marker}. Our method achieves an accuracy of 92.59\% in this task, with only two slip cases mis-recognized. For grasping experiments, a robotic gripper is programmed to lift an object. Sensor WO and WM are positioned on either side of the gripper, where Sensor WM remains inactive and serves solely as a physical support to ensure the contact points on the both sides hold at the same height. The gripping force was set to just the right amount to make slip happens. The slip detection algorithm serves as feedback to enable gripper control for promptly halting slip. In our experiment, the robot gripper ascends at a rate of 6mm/s, while Sensor WO captures data at 30Hz. A successful trial requires the accurate slip detection and the stable lifting of the object by the robot gripper. We used five objects shown in Fig.~\ref{Fig.grasp-obj-demo}, and all of them were successfully lifted by the gripper.
 
 A group of qualitative results is showcased in Fig.~\ref{Fig.grasp-vitac}. At the first frame, the initial motions of markers are recorded as reference. Next, our algorithm decides if the slip condition (\ref{Eq.slip-condition}) satisfies for each subsequent frame. Afterwards, when slip is successfully detected (the 2$^\textup{nd}$ column), the robot gripper is controlled to increase grasping force to halt slip, as shown in the 2$^\textup{nd}$ and 3$^\textup{rd}$ columns. Finally, the object is successfully lifted, with the corresponding tactile signal almost unchanged (the 4$^\textup{th}$ column).

\section{Conclusion}
\label{Sec.Conclusion}
 In this paper, we propose a mode-switchable optical tactile sensing approach to carry out bidirectional marker-markerless transitions. Experiments show that our method has the potential to facilitate both perception and manipulation tasks. In the future study, we plan to eliminate reliance on data collection and leverage simulation tools~\cite{Daniel-sim2real,Sim-to-real-texture-generation} to train the models and investigate the sim2real capability of our method. Furthermore, we will extend the application of our method to more dexterous manipulation tasks like in-hand rotation.

 \section{Acknowledgments}
We thank Prof. Edward Adelson at Massachusetts Institute of Technology for engaging discussions on inpainting markers and for sharing insights into mode-switchable optical tactile sensing.
	\bibliographystyle{IEEEtran}
	\bibliography{ref2}\ 

\begin{thebibliography}{10}
\providecommand{\url}[1]{#1}
\csname url@rmstyle\endcsname
\providecommand{\newblock}{\relax}
\providecommand{\bibinfo}[2]{#2}
\providecommand\BIBentrySTDinterwordspacing{\spaceskip=0pt\relax}
\providecommand\BIBentryALTinterwordstretchfactor{4}
\providecommand\BIBentryALTinterwordspacing{\spaceskip=\fontdimen2\font plus
\BIBentryALTinterwordstretchfactor\fontdimen3\font minus
  \fontdimen4\font\relax}
\providecommand\BIBforeignlanguage[2]{{%
\expandafter\ifx\csname l@#1\endcsname\relax
\typeout{** WARNING: IEEEtran.bst: No hyphenation pattern has been}%
\typeout{** loaded for the language `#1'. Using the pattern for}%
\typeout{** the default language instead.}%
\else
\language=\csname l@#1\endcsname
\fi
#2}}

\bibitem{luo2017-tactile-perception-review}
S.~Luo, J.~Bimbo, R.~Dahiya, and et~al., ``Robotic tactile perception of object
  properties: A review,'' \emph{Mechatronics}, vol.~48, pp. 54--67, 2017.

\bibitem{Gelsight}
W.~Yuan, S.~Dong, and E.~H. Adelson, ``{GelSight}: {High-Resolution} robot
  tactile sensors for estimating geometry and force,'' \emph{Sensors}, vol.~17,
  no.~12, p. 2762, 2017.

\bibitem{Gelslim}
E.~Donlon, S.~Dong, M.~Liu, and et~al., ``{GelSlim}: A high-resolution,
  compact, robust, and calibrated tactile-sensing finger,'' in \emph{IROS},
  2018, pp. 1927--1934.

\bibitem{Tactip-family}
B.~Ward-Cherrier, N.~Pestell, L.~Cramphorn, and et~al., ``The {TacTip} family:
  Soft optical tactile sensors with 3d-printed biomimetic morphologies,''
  \emph{Soft robotics}, vol.~5, no.~2, pp. 216--227, 2018.

\bibitem{Measure-shear-slip}
W.~Yuan, R.~Li, M.~A. Srinivasan, and et~al., ``Measurement of shear and slip
  with a {GelSight} tactile sensor,'' in \emph{ICRA}, 2015, pp. 304--311.

\bibitem{Gelslim-V2}
D.~Ma, E.~Donlon, S.~Dong, and et~al., ``Dense tactile force estimation using
  {GelSlim} and inverse fem,'' in \emph{ICRA}, 2019, pp. 5418--5424.

\bibitem{improved-gelsight}
S.~Dong, W.~Yuan, and E.~H. Adelson, ``Improved {Gelsight} tactile sensor for
  measuring geometry and slip,'' in \emph{IROS}, 2017, pp. 137--144.

\bibitem{Gripper-control1}
J.~Ueda, A.~Ikeda, and T.~Ogasawara, ``Grip-force control of an elastic object
  by vision-based slip-margin feedback during the incipient slip,'' \emph{IEEE
  Transactions on Robotics}, vol.~21, no.~6, pp. 1139--1147, 2005.

\bibitem{Gripper-control2}
S.~Denei, P.~Maiolino, E.~Baglini, and et~al., ``Development of an integrated
  tactile sensor system for clothes manipulation and classification using
  industrial grippers,'' \emph{IEEE Sensors Journal}, vol.~17, no.~19, pp.
  6385--6396, 2017.

\bibitem{Active-clothing-perception}
W.~Yuan, Y.~Mo, S.~Wang, and et~al., ``Active clothing material perception
  using tactile sensing and deep learning,'' in \emph{ICRA}, 2018, pp.
  4842--4849.

\bibitem{Spatio-temporal-attention}
G.~Cao, Y.~Zhou, D.~Bollegala, and et~al., ``Spatio-temporal attention model
  for tactile texture recognition,'' in \emph{IROS}, 2020, pp. 9896--9902.

\bibitem{GelFinger}
Z.~Lin, J.~Zhuang, Y.~Li, and et~al., ``{GelFinger}: A novel visual-tactile
  sensor with multi-angle tactile image stitching,'' \emph{RA-L}, 2023.

\bibitem{Low-cost-GelSight-UV}
A.~C. Abad and A.~Ranasinghe, ``Low-cost {GelSight} with uv markings: Feature
  extraction of objects using alexnet and optical flow without 3d image
  reconstruction,'' in \emph{ICRA}, 2020, pp. 3680--3685.

\bibitem{UVtac}
W.~Kim, W.~D. Kim, J.-J. Kim, and et~al., ``{UVtac}: Switchable uv marker-based
  tactile sensing finger for effective force estimation and object
  localization,'' \emph{RA-L}, vol.~7, no.~3, pp. 6036--6043, 2022.

\bibitem{Tactip-original}
C.~Chorley, C.~Melhuish, T.~Pipe, and et~al., ``Development of a tactile sensor
  based on biologically inspired edge encoding,'' in \emph{International
  Conference on Advanced Robotics}, 2009, pp. 1--6.

\bibitem{TacCylinder}
B.~Winstone, C.~Melhuish, T.~Pipe, and et~al., ``Toward bio-inspired tactile
  sensing capsule endoscopy for detection of submucosal tumors,'' \emph{IEEE
  Sensors Journal}, vol.~17, no.~3, pp. 848--857, 2016.

\bibitem{TacThumb}
B.~Ward-Cherrier, L.~Cramphorn, and N.~F. Lepora, ``Tactile manipulation with a
  tacthumb integrated on the open-hand m2 gripper,'' \emph{RA-L}, vol.~1,
  no.~1, pp. 169--175, 2016.

\bibitem{GelSight-original}
R.~Li, R.~Platt, W.~Yuan, and et~al., ``Localization and manipulation of small
  parts using gelsight tactile sensing,'' in \emph{IROS}, 2014, pp. 3988--3993.

\bibitem{GelTip}
D.~F. Gomes, Z.~Lin, and S.~Luo, ``{GelTip}: A finger-shaped optical tactile
  sensor for robotic manipulation,'' in \emph{IROS}, 2020, pp. 9903--9909.

\bibitem{GAN-original}
I.~Goodfellow, J.~Pouget-Abadie, M.~Mirza, and et~al., ``Generative adversarial
  nets,'' \emph{NeurIPS}, vol.~27, 2014.

\bibitem{VAE-original}
D.~P. Kingma and M.~Welling, ``Auto-encoding variational bayes,'' \emph{arXiv
  preprint arXiv:1312.6114}, 2013.

\bibitem{Deep-autoregressive-Networks}
K.~Gregor, I.~Danihelka, A.~Mnih, and et~al., ``Deep autoregressive networks,''
  in \emph{ICML}, 2014, pp. 1242--1250.

\bibitem{Normalizing-Flows1}
D.~Rezende and S.~Mohamed, ``Variational inference with normalizing flows,'' in
  \emph{ICML}, 2015, pp. 1530--1538.

\bibitem{DDPM}
J.~Ho, A.~Jain, and P.~Abbeel, ``Denoising diffusion probabilistic models,''
  \emph{NeurIPS}, vol.~33, pp. 6840--6851, 2020.

\bibitem{Sim-to-real-CycleGAN}
W.~Chen, Y.~Xu, Z.~Chen, and et~al., ``Bidirectional sim-to-real transfer for
  {GelSight} tactile sensors with cyclegan,'' \emph{RA-L}, vol.~7, no.~3, pp.
  6187--6194, 2022.

\bibitem{Sim-to-real-ACTNet}
X.~Jing, K.~Qian, T.~Jianu, and et~al., ``Unsupervised adversarial domain
  adaptation for sim-to-real transfer of tactile images,'' \emph{IEEE
  Transactions on Instrumentation and Measurement}, 2023.

\bibitem{Sim-to-real-texture-generation}
T.~Jianu, D.~F. Gomes, and S.~Luo, ``Reducing tactile sim2real domain gaps via
  deep texture generation networks,'' in \emph{ICRA}, 2022, pp. 8305--8311.

\bibitem{Touching-to-see}
J.-T. Lee, D.~Bollegala, and S.~Luo, ``{`Touching to see' and `seeing to feel':
  Robotic cross-modal sensory data generation for visual-tactile perception},''
  in \emph{ICRA}, 2019, pp. 4276--4282.

\bibitem{cao2023vis2hap}
G.~Cao, J.~Jiang, N.~Mao, D.~Bollegala, and et~al., ``{Vis2Hap}: Vision-based
  haptic rendering by cross-modal generation,'' in \emph{ICRA}, 2023, pp.
  12\,443--12\,449.

\bibitem{TacMAE}
G.~Cao, J.~Jiang, D.~Bollegala, and et~al., ``Learn from incomplete tactile
  data: Tactile representation learning with masked autoencoders,'' in
  \emph{IROS}, 2023, pp. 10\,800--10\,805.

\bibitem{Palette}
C.~Saharia, W.~Chan, H.~Chang, and et~al., ``Palette: Image-to-image diffusion
  models,'' in \emph{ACM SIGGRAPH}, 2022, pp. 1--10.

\bibitem{UNet}
O.~Ronneberger, P.~Fischer, and T.~Brox, ``{U-Net}: Convolutional networks for
  biomedical image segmentation,'' in \emph{MICCAI}.\hskip 1em plus 0.5em minus
  0.4em\relax Springer, 2015, pp. 234--241.

\bibitem{TELEA}
A.~Telea, ``An image inpainting technique based on the fast marching method,''
  \emph{Journal of graphics tools}, vol.~9, no.~1, pp. 23--34, 2004.

\bibitem{NS}
M.~Bertalmio, A.~L. Bertozzi, and G.~Sapiro, ``Navier-stokes, fluid dynamics,
  and image and video inpainting,'' in \emph{CVPR}, vol.~1, 2001, pp. I--I.

\bibitem{FMM}
J.~A. Sethian, ``A fast marching level set method for monotonically advancing
  fronts.'' \emph{Proceedings of the National Academy of Sciences}, vol.~93,
  no.~4, pp. 1591--1595, 1996.

\bibitem{DeepLabV3}
L.-C. Chen, G.~Papandreou, F.~Schroff, and et~al., ``Rethinking atrous
  convolution for semantic image segmentation,'' \emph{arXiv preprint
  arXiv:1706.05587}, 2017.

\bibitem{GelWedge}
S.~Wang, Y.~She, B.~Romero, and E.~Adelson, ``{GelSight Wedge}: Measuring
  high-resolution 3d contact geometry with a compact robot finger,'' in
  \emph{ICRA}, 2021, pp. 6468--6475.

\bibitem{pix2pix}
P.~Isola, J.-Y. Zhu, T.~Zhou, and et~al., ``Image-to-image translation with
  conditional adversarial networks,'' in \emph{CVPR}, 2017, pp. 1125--1134.

\bibitem{Daniel-sim2real}
D.~F. Gomes, P.~Paoletti, and S.~Luo, ``Generation of {GelSight} tactile images
  for sim2real learning,'' \emph{RA-L}, vol.~6, no.~2, pp. 4177--4184, 2021.

\bibitem{Vitac}
S.~Luo, W.~Yuan, E.~Adelson, and et~al., ``{ViTac}: Feature sharing between
  vision and tactile sensing for cloth texture recognition,'' in \emph{ICRA},
  2018, pp. 2722--2727.

\bibitem{Cold-diffusion}
A.~Bansal, E.~Borgnia, H.-M. Chu, and et~al., ``Cold diffusion: Inverting
  arbitrary image transforms without noise,'' \emph{NeurIPS}, vol.~36, 2024.

\bibitem{Marker-embedded-image-generation}
W.~D. Kim, S.~Yang, W.~Kim, and et~al., ``{Marker-Embedded} tactile image
  generation via generative adversarial networks,'' \emph{RA-L}, 2023.

\bibitem{ResNet}
K.~He, X.~Zhang, S.~Ren, and et~al., ``Deep residual learning for image
  recognition,'' in \emph{CVPR}, 2016, pp. 770--778.

\end{thebibliography}
	%
	%
	%
	
\end{document}